\documentclass{article}

\usepackage{arxiv}

\usepackage[utf8]{inputenc} 
\usepackage[T1]{fontenc}    
\usepackage{hyperref}       
\usepackage{url}            
\usepackage{booktabs}       
\usepackage{amsfonts}       
\usepackage{nicefrac}       
\usepackage{microtype}      
\usepackage{lipsum}
\usepackage{amsmath}   
\usepackage[table]{xcolor}    
\usepackage{enumitem}  
\usepackage{float}  
\usepackage{graphicx}
\graphicspath{ {./images/} }

\title{CAR3T: Generalizable Calibration-Guided Reference-View Attention Reuse for Multi-View 3D Transformers}

\author{
\shortstack[c]{
{\normalfont\normalsize\upshape
Lan Wei$^{1}$,
Haosong Peng$^{2}$,
Sicheng Tian$^{1}$,
Tianyu Qi$^{3}$,
Wenchao Xu$^{2}$,
Lei Xue$^{3}$,
Yufeng Zhan$^{1}$,
Yuanqing Xia$^{1}$
}\\[0.4em]
{\normalfont\small\upshape
$^{1}$ Beijing Institute of Technology
\qquad
$^{2}$ The Hong Kong University of Science and Technology
}\\[0.2em]
{\normalfont\small\upshape
$^{3}$ SUN YAT-SEN UNIVERSITY
}
}
}

\begin{document}
\maketitle
\begin{abstract}
Feed-forward multi-view 3D reconstruction relies on global attention for cross-view aggregation, but the attention cost grows quadratically with the number of views, making inference increasingly expensive for larger inputs. Existing acceleration methods compress tokens or sparsify attention, potentially losing geometric detail or restricting cross-view aggregation while leaving redundancy among attention outputs underexploited.
We observe similar cross-view attention distributions in certain layers, revealing an opportunity to reduce repeated global aggregation through output reuse. However, the layers suitable for reuse vary with the input, and reusing one layer alters the features received by later layers. Similarity at individual layers therefore cannot characterize the combined effect of multiple reuse decisions on output drift. 
To address these challenges, we propose CAR3T, a training-free framework combining reference-view attention output reuse with offline risk calibration and online dynamic layer selection. Calibration models cumulative reuse risk, while online selection adapts reuse to each input under a calibrated budget. CAR3T retains all input tokens and attends to the keys and values of all views, reducing the core attention computation in selected layers from quadratic to linear. Experiments on VGGT, DA3, and $\pi^3$ show forward speedups up to $1.41\times$ with at most 100 views and $1.75\times$ on longer sequences, with accuracy trade-offs across backbones and datasets.
\end{abstract}

\section{Introduction}
\label{sec:Introduction}

Multi-view 3D reconstruction recovers scene geometry and camera parameters from images, supporting robotic perception~\cite{omnivggt,peng2026spatialbenchspatialfoundationmodel} and autonomous driving~\cite{dvgt,yuan2026surroundnexo}. 
Current feed-forward models such as VGGT~\cite{wang2025vggt}, Depth Anything 3 (DA3)~\cite{lin2025da3}, and $\pi^3$~\cite{wang2026pi3} learn cross-view feature interactions and predict geometry and camera parameters in a single forward pass. They aggregate information across views through global attention, whose efficiency is central to scaling reconstruction to larger inputs.

However, global attention remains a bottleneck when scaling these models to many views. Each view contributes queries that attend to the keys and values of all views, so the core attention computation grows quadratically with the number of input views. Efficient kernels such as FlashAttention reduce memory access costs~\cite{dao2022flashattention}, but retain the quadratic computation of exact dense attention. With dozens or hundreds of input views, repeating global aggregation for every query becomes expensive.
 
Existing acceleration methods reduce this cost through token compression~\cite{shen2025fastvggt,shu2026litevggt,chen2026come}, sparse attention~\cite{sun2026avggt,deng2026saf3r}, or compact global representations~\cite{wang2026flashvggt}. Such compression or sparsification can lose geometric detail or limit information exchange, yet redundancy among attention outputs remains underexploited. To examine this possibility, we analyze cross-view attention in VGGT, DA3, and $\pi^3$ and observe similar attention distributions in certain layers. Since queries across views share the same global keys and values, similar distributions can produce similar aggregation responses. This observation motivates reusing a reference view's attention response across views to reduce repeated computation while preserving all input tokens and access to all keys and values.

However, applying this reuse poses challenges in both layer selection and the control of output drift. Cross-view attention similarity varies across layers and inputs, so the suitability of reuse changes with the current features. Moreover, reusing one layer alters the features passed to later layers, making the final output drift depend on the combination of reused layers. Similarity at one layer therefore provides an incomplete basis for multiple reuse decisions. Effective reuse requires adapting layer selection to the input while accounting for cumulative output drift and interactions among the selected layers.

To address these challenges, we propose CAR3T, a training-free framework that combines reference-view attention output reuse with offline risk calibration and online dynamic layer selection. In each selected layer, CAR3T computes the reference view's attention response over the keys and values of all views and reuses it across views. With one reference view, this reduces the core attention computation from quadratic to linear in the number of views. The query, key, value, and subsequent output projections remain unchanged. Other layers retain the original global attention to preserve view-specific aggregation. Offline calibration relates Query Drift to Output Drift and models single-layer sensitivity, pairwise interactions, and higher-order combination risks. A Pareto analysis of latency and output drift on validation inputs determines the reuse budget. During inference, the online module combines query features, cumulative risk, and a lookahead estimate of future reuse value to guide layer selection under calibrated risk constraints. CAR3T operates with unchanged pre-trained weights and requires no fine-tuning. Experiments on VGGT, DA3, and $\pi^3$ across eight benchmark settings show overall forward speedups of $1.13\times$ to $1.31\times$ with up to 100 input views. Longer sequences yield speedups up to $1.75\times$. Pose and reconstruction evaluations characterize the resulting accuracy trade-offs across backbones and input scales. Our main contributions are:
\begin{itemize}[leftmargin=2em]
    \item We introduce reference-view attention output reuse that reduces the core attention computation while retaining all input tokens and attending to all keys and values.
    \item We combine offline risk calibration with online dynamic layer selection to adapt reuse to each input under a calibrated budget, accounting for cumulative risk and interactions among reused layers.
    \item We evaluate CAR3T on three feedforward reconstruction models, showing inference speedups across input scales and quantifying the geometric accuracy trade-offs.
\end{itemize}
\section{Related Work}
\label{sec:Related Work}


\noindent\textbf{Feedforward Visual Geometry Models.}
In recent years, feedforward visual geometry models have gradually emerged as a new paradigm distinct from traditional multi-stage optimization pipelines.
DUSt3R~\cite{wang2024dust3r} regresses pixel-aligned 3D point maps from uncalibrated image pairs, while MASt3R~\cite{leroy2024mast3r} extends it with local features for robust dense matching.
MV-DUSt3R+~\cite{MV-DUSt3R+}, Fast3R~\cite{yang2025fast3r}, MUSt3R~\cite{must3r}, MVSplat~\cite{mvsplat}, and DepthSplat~\cite{DepthSplat} have advanced multi-view 3D reconstruction using point maps and 3D Gaussians.
With further advancements, VGGT~\cite{wang2025vggt} unifies multi-view visual geometry modeling using a feedforward Transformer architecture, simultaneously predicting camera parameters, depth maps, point maps, and 3D point trajectories through alternating frame-wise and global attention mechanisms.  
$\pi^3$~\cite{wang2026pi3} and DA3~\cite{lin2025da3} further reduce reliance on fixed reference frames and specialized architectures, collectively advancing feedforward 3D reconstruction toward unified and generalizable multi-view geometry models. 
However, these models fundamentally depend on dense cross-view global feature interactions, leading to a rapid increase in computational cost with growing number of views, which becomes a significant efficiency bottleneck when scaling to long sequences and large-scale scenes.

\noindent\textbf{Efficient Inference for Feedforward 3D Reconstruction Models.}
Efficient feedforward reconstruction methods reduce computation through \emph{1) token compression}, \emph{2) compact scene representations}, or \emph{3) attention sparsification}. 
FastVGGT~\cite{shen2025fastvggt} prunes tokens based on aggregation patterns, while LiteVGGT~\cite{shu2026litevggt}, HTTM~\cite{wang2026httm}, and Co-Me~\cite{chen2026come} use geometry-aware caching-based, head-wise temporal, and confidence-guided merging, respectively. Spark3R~\cite{tang2026spark3rasymmetrictokenreduction} combines asymmetric query merging, key-value pruning, and layer-adaptive scheduling. Such compression reduces attention cost, but merged or removed tokens no longer contribute independently, and aggressive compression may lose view-specific correspondences and geometric detail.
For \emph{compact scene representations}, FlashVGGT~\cite{wang2026flashvggt} aggregates information through cross-attention between frame descriptors and image tokens, while VGG-T\(^{3}\)~\cite{elflein2026vggt3} uses test-time training to distill variable-length scenes into a fixed-size state. These approaches trade representation capacity against geometric fidelity.
For \emph{attention sparsification}, Block-Sparse Global Attention~\cite{wang2026blocksparse} replaces dense cross-view interactions with block-sparse connections, AVGGT~\cite{sun2026avggt} uses frame-wise attention and sampled key-value tokens, and SAF3R~\cite{deng2026saf3r} adaptively selects head-specific computation modes. Reducing these connections can limit cross-view aggregation and does not explicitly model redundancy among attention outputs across views.


\section{Methodology}
\label{sec:Methodology}

This section introduces CAR3T, a framework for accelerating inference in multi-view 3D Transformers. 
We first outline the overall workflow of CAR3T in Sec.~\ref{sec:4.1}. We then motivate global attention reuse through an analysis of cross-view attention similarity and define the reference query reuse operator in Sec.~\ref{sec:4.2}. Sec.~\ref{sec:4.3} describes the offline calibration of reuse risk and budgets. Finally, Sec.~\ref{sec:4.4} presents calibration-guided online layer selection.

\begin{figure*}[t]
    \centering
    \includegraphics[width=1\textwidth]{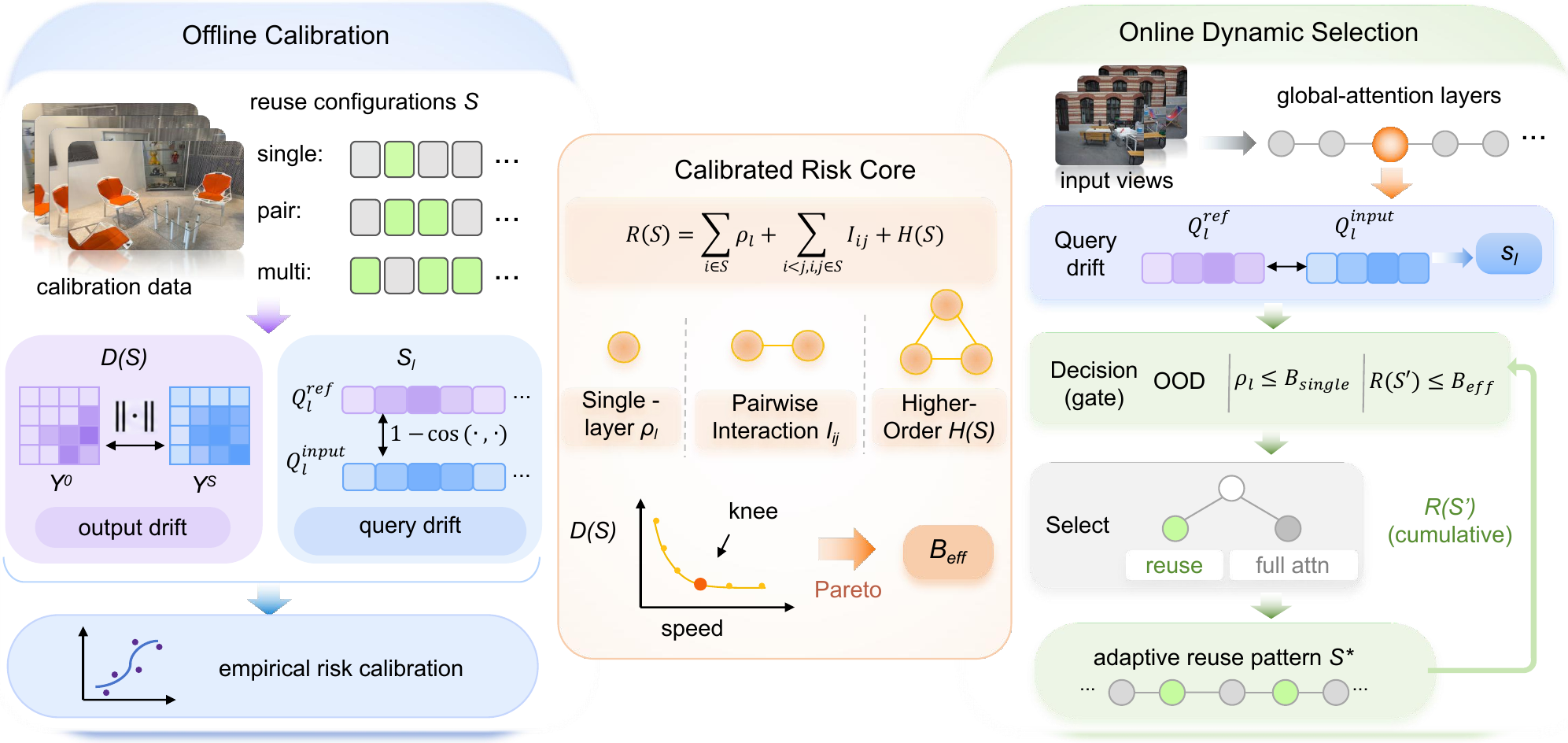}
    \caption{\textbf{Overview of CAR3T.} Offline calibration establishes a risk model accounting for single-layer, pairwise, and higher-order interactions, with the effective risk budget determined by the speed–drift Pareto frontier. 
    During inference, online query drift and cumulative risk constraints guide input-dependent reuse-layer selection.}
    \label{fig:2}
    \vspace{-3mm}
\end{figure*}

\subsection{Overview of CAR3T}
\label{sec:4.1}

As shown in Fig.~\ref{fig:2}, CAR3T adopts a framework combining \emph{offline} risk calibration with \emph{online} dynamic layer selection to selectively reuse multi-view global attention, thereby reducing computational overhead and controlling model output drift caused by reuse. 

Let $\mathcal{L}$ denote the set of indices for candidate global attention layers. 
In the \emph{offline} stage, we evaluate reuse configurations $\mathcal{S} \subseteq \mathcal{L}$ in terms of output drift $D(\mathcal{S})$ and inference latency $T(\mathcal{S})$, and calibrate an empirical risk model $\widehat{\mathcal{R}}$. 
In the \emph{online} stage, given multi-view input $\mathcal{I} = \{I_i\}_{i=1}^{M}$ with $M$ views and the calibrated risk budget $B_{\mathrm{eff}}$, we use the risk model $\widehat{\mathcal{R}}$ to dynamically select the reuse layers $\mathcal{S}^{*}(\mathcal{I}) \subseteq \mathcal{L}$.

For each layer $\ell \in \mathcal{S}^{*}(\mathcal{I})$, CAR3T reuses attention outputs from a certrain reference view $r \in \{1,\ldots,M\}$ across views; for all other layers, original global attention computations are preserved. 
Without fine-tuning the pre-trained model or training additional selection networks, CAR3T dynamically adjusts its reuse strategy according to input, achieving a controllable trade-off between inference efficiency and model output consistency.

\subsection{Motivation of Global Attention Reuse}
\label{sec:4.2}
\begin{figure*}[t]
    \centering
    \includegraphics[
        width=1\textwidth,
        trim=0cm 0cm 0cm 0cm,
        clip
    ]{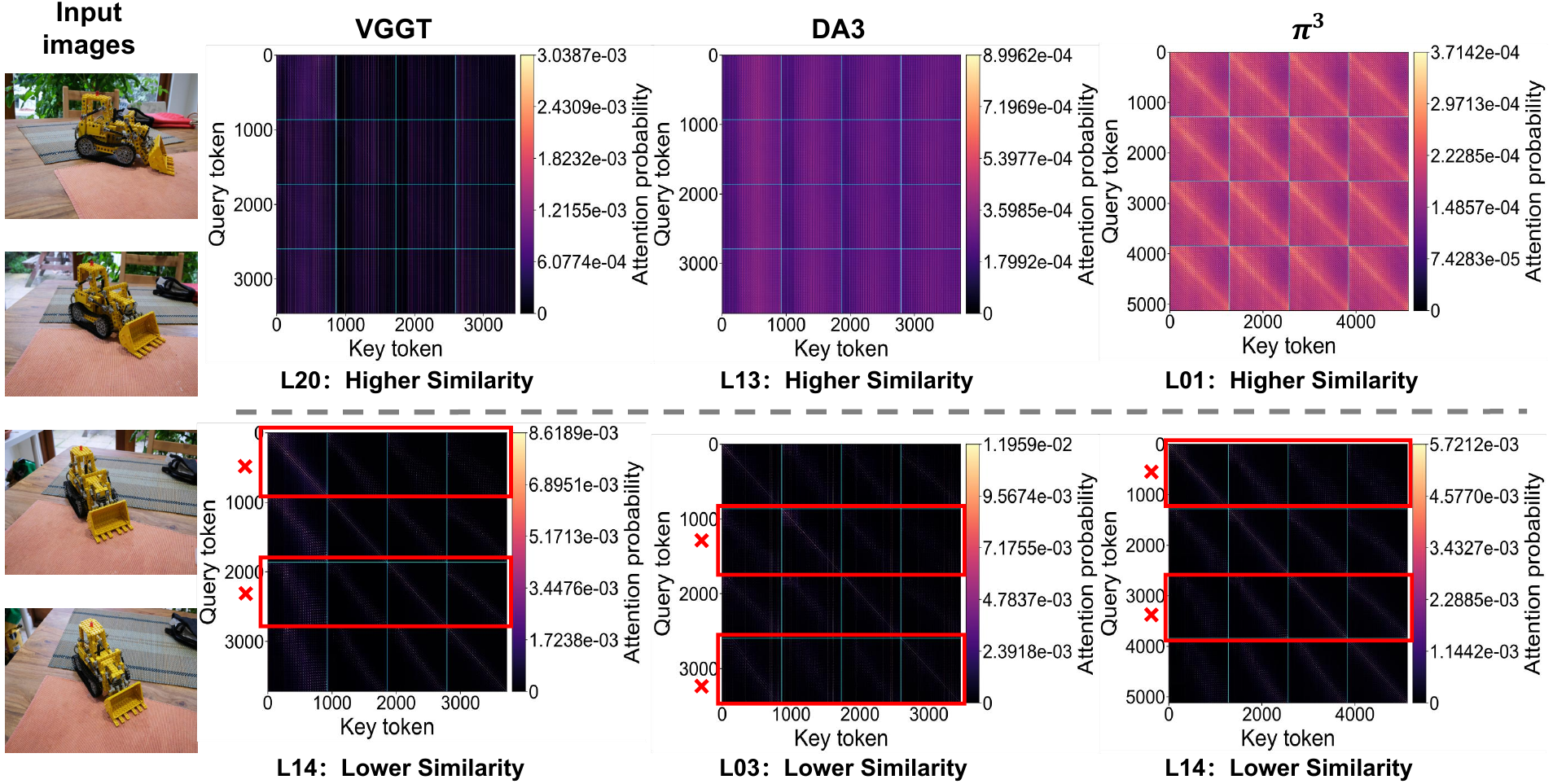}
    \caption{\textbf{Cross-view attention in VGGT, DA3, and $\pi^3$.} Left: input image. Right: representative attention maps at different layers for each model. The top and bottom rows show layers with higher and lower cross-view similarity, respectively, with red boxes and crosses highlighting regions with large discrepancies in attention distribution.}
    \label{fig:1}
\end{figure*}

\begin{figure*}[!htbp]
    \centering
    \includegraphics[width=1\textwidth,]{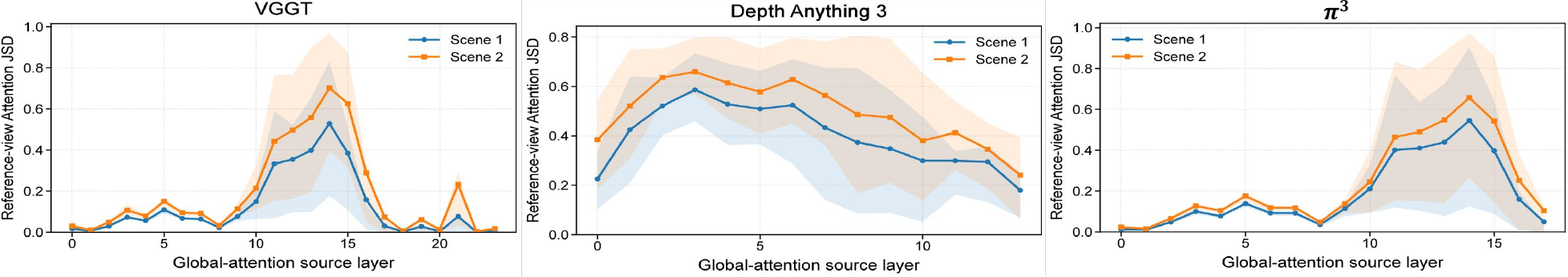}
    \caption{\textbf{Differences in cross-view attention distributions across global attention layers among VGGT, \(\pi^3\), and Depth Anything 3. } 
    A lower JSD indicates smaller differences and higher similarity between the two attention distributions.}
    \label{fig:3}
      \vspace{-3mm}
\end{figure*}

\noindent\textbf{Layer-wise Differences in Cross-view Attention.} 
To study how global attention varies across views and layers, we visualize the attention maps of VGGT, DA3, and $\pi^3$. 
As shown in Fig.~\ref{fig:1}, we feed the same four images to each model. 
Across all three models, attention block rows corresponding to different query views show similar patterns in some layers, but are less similar in others.
To quantify this similarity, we compute the Jensen--Shannon divergence (JSD)~\cite{lin1991divergence} between the global attention distributions of corresponding query tokens in the reference view and the other views. A lower JSD indicates more similar distributions. We average the scores across tokens and non-reference views for each attention head.
Appx~\ref{app:jsd} provides the calculation details.
As shown in Fig.~\ref{fig:3}, the solid lines show the mean JSD across attention heads, and the shaded regions show the interquartile range. 
The JSD varies across layers in all three models, with lower values in some layers. 
Together, these qualitative and quantitative results suggest that cross-view attention can be redundant, but that the degree of redundancy varies across layers. This observation motivates selectively reusing the attention response of a reference view across views in suitable layers.

\noindent\textbf{Definition of Reference Query Reuse Operator.} 
To exploit this redundancy, we define a reference query reuse operator. Let $P$ denote the number of tokens per view and $d$ the feature dimension of a single attention head. At layer $\ell$, the operator computes the attention response using the queries $\mathbf Q_r^{(\ell)}$ of a reference view $r$ and the concatenated keys $\mathbf K^{(\ell)}$ and values $\mathbf V^{(\ell)}$ of all views:

\begin{equation}
\mathbf O_r^{(\ell)}
=
\operatorname{Softmax}
\left(
\frac{
\mathbf Q_r^{(\ell)}
\mathbf K^{(\ell)\top}
}{\sqrt{d}}
\right)
\mathbf V^{(\ell)}.
\end{equation}

The response is then reused for all views at the same level:

\begin{equation}
\widetilde{\mathbf O}_i^{(\ell)}
=
\mathbf O_r^{(\ell)},
\qquad i=1,\ldots,M.
\end{equation}

This reduces the number of queries in the attention computation from $MP$ to $P$, while retaining all $MP$ keys and values. Ignoring batch size and the number of attention heads, the complexity of the attention core is therefore reduced from $\mathcal{O}(M^2P^2d)$ to $\mathcal{O}(MP^2d)$ (see Appx.~\ref{sec:Preliminary}). The Q/K/V projections and subsequent output projections for all views are still performed as usual.

\subsection{Offline Calibration of Reuse Risk and Budgets}
\label{sec:4.3}

CAR3T uses offline calibration to guide online layer selection.
The offline stage has two objectives: first, to establish an empirical mapping between \emph{Output Drift} and \emph{Query Drift}, analyze the impact of different layer reuse configurations on model outputs, and thereby build a unified reuse risk model; second, to evaluate the inference efficiency and output quality of different static reuse configurations on the validation set, select a trade-off configuration through Pareto analysis, and use it to determine the risk budget required for online inference.

\noindent\textbf{Definition of Output Drift.} 
To quantify the effect of reuse on model outputs, we consider three output components: \emph{1) per-view geometry (depth or local points)}, \emph{2) global points}, and \emph{3) camera poses}. For a set of reused layers $\mathcal{S}$, let $Y_k^{\mathcal{S}}$ and $Y_k^0$ denote the $k$-th ($k\in\{1,2,3\}$) component produced with reuse and by the full model, respectively. We define the relative drift of each component and their equally weighted mean as:

\begin{equation}
D_k(\mathcal{S})=
\frac{\left\|Y_k^{\mathcal{S}}-Y_k^0\right\|_2}
{\left\|Y_k^0\right\|_2+\epsilon},
\quad
D(\mathcal{S})=\frac{1}{3}\sum_{k=1}^{3}D_k(\mathcal{S}).
\tag{6}
\end{equation}

\paragraph{Query Drift Estimation.}

We use per-layer query drift as an input-dependent proxy, calibrating its relationship with output drift through offline ablation studies.
Specifically, let $r$ denote the reference view and $v$ denote another view ($v\neq r$). At layer $\ell$, let $Q_{r,h,n}^{(\ell)}$ and $Q_{v,h,n}^{(\ell)}$ denote the query vectors of views $r$ and $v$, respectively, at the same token position $n$ and attention head $h$. Here, $\ell$, $h$, and $n$ represent the indices of the network layer, attention head, and token position, respectively.

We measure the difference between the two query vectors using cosine distance:

\begin{equation}
\delta_{v,h,n}^{(\ell)}
=
1-\cos\left(
Q_{r,h,n}^{(\ell)},
Q_{v,h,n}^{(\ell)}
\right).
\tag{8}
\end{equation}

We summarize query differences at each layer with the query drift score:

\begin{equation}
s_\ell
=
m_\ell+0.5p_\ell+0.25x_\ell.
\tag{9}
\end{equation}
Here, $m_\ell$, $p_\ell$, and $x_\ell$ are the mean, 90th percentile, and maximum of the distances across all sampled queries, respectively.

\noindent\textbf{Single-layer Risk Calibration.} 
We profile each layer offline to calibrate the relationship between query drift and output drift. During inference, we estimate the risk of reusing layer $\ell$ as:

\begin{equation}
\rho_\ell =
u_\ell \frac{s_\ell}{\bar{s}_\ell}.
\tag{10}
\end{equation}
Here, $\bar{s}_\ell$ is the average query drift measured offline, and $u_\ell$ is a conservative sensitivity coefficient derived from a high quantile of the offline risk ratio.

\noindent\textbf{Two-layer and Higher-order Risk Modeling.} 
Beyond single-layer risks, we account for two-layer interactions and higher-order effects when reusing multiple layers.
Define the offline interaction residual for layer pair $(i,j)$ as:

\begin{equation}
I_{ij}
=
D(\{i,j\})-D(\{i\})-D(\{j\}).
\tag{11}
\end{equation}

For the two-layer test results, CAR3T adopts a relatively high interaction risk value based on the offline experimental results to avoid underestimating the error.
For combinations with more reused layers, it further leverages offline combination ablation to build a higher-order risk model $H(\mathcal{S})$ to capture risks that cannot be explained by single layers and pairwise interactions. Thus, the overall risk of a set of reused layers is expressed as:

\begin{equation}
R(\mathcal{S})=
\sum_{\ell\in\mathcal{S}} \rho_\ell
+\sum_{\substack{i<j\\i,j\in\mathcal{S}}} I_{ij}
+H(\mathcal{S}).
\tag{12}
\end{equation}

The model maps the current input query features to the empirical risk of a set of reusable layers, providing a unified risk metric for subsequent dynamic decision-making.

\noindent\textbf{Pareto-based Optimal Configuration Selection.}
We evaluate the relative speedup gains of different static reuse configurations on the validation set:
\begin{equation}
G(\mathcal{S})
=
\max\left(
0,
\frac{t_0-t_{\mathcal{S}}}{t_0}
\right),
\tag{13}
\end{equation}
where $t_0$ and $t_{\mathcal{S}}$ denote the inference latency of the full model and the reused configuration, respectively.
By incorporating the output drift $D(\mathcal{S})$, we construct the quality–efficiency Pareto frontier and select knee-point candidates based on the normalized speedup and output drift:

\begin{equation}
K(\mathcal{S})
=
\widetilde{G}(\mathcal{S})
-\lambda_D\widetilde{D}(\mathcal{S}).
\tag{14}
\end{equation}

Among them, $\lambda_D$ controls the trade-off between output consistency and computational efficiency. 

\noindent\textbf{Risk Budget Determination.} 
After selecting the elbow-point configuration, we re-evaluate the set of reusable layers using the recorded query features and the same risk model used during online inference. The resulting total risk serves as the reference budget $B_{\mathrm{eff}}$ for the effective cumulative risk, ensuring that budget calibration and online selection use the same risk metric. The individual single-layer limit $B_{\mathrm{single}}$ prevents any single layer from consuming too much budget. We calibrate separately for different ranges of the number of views and select the corresponding calibration result based on the number of input views.

\subsection{Calibration-Guided Online Layer Selection}
\label{sec:4.4}

\noindent\textbf{Online Dynamic Layer Selection.} 
During inference, CAR3T decides, in network layer order, whether to reuse each global attention layer. At layer $\ell$, let $\mathcal{S}_{\ell-1}$ denote the set of layers selected so far, and let $\mathcal{S}'=\mathcal{S}_{\ell-1}\cup\{\ell\}$. CAR3T first checks whether the number of input views and the query drift are within the calibrated range. It then performs the standard checks for the single-layer risk constraint and the cumulative risk constraint:

\begin{equation}
\rho_\ell\leq B_{\mathrm{single}},
\qquad
R(\mathcal{S}')\leq B_{\mathrm{eff}}.
\tag{15}
\end{equation}

To reserve budget for subsequent layers, CAR3T compares two options: reusing the current layer and consuming part of the remaining budget, or skipping the current layer and retaining the budget. For each option, approximate dynamic programming uses offline risk estimates to predict how many subsequent layers can be reused within the remaining budget. CAR3T prefers the option that is expected to yield a larger total number of reusable layers, counting the current layer toward the total if it is reused. When it reaches a subsequent layer, CAR3T re-evaluates and makes a decision using the layer’s actual query features.

\section{Experiments}
\label{sec:Experiments}

\subsection{Experimental Setup}
\label{sec:5.1}
\begin{figure*}[t]
    \centering
    \includegraphics[width=1\textwidth]{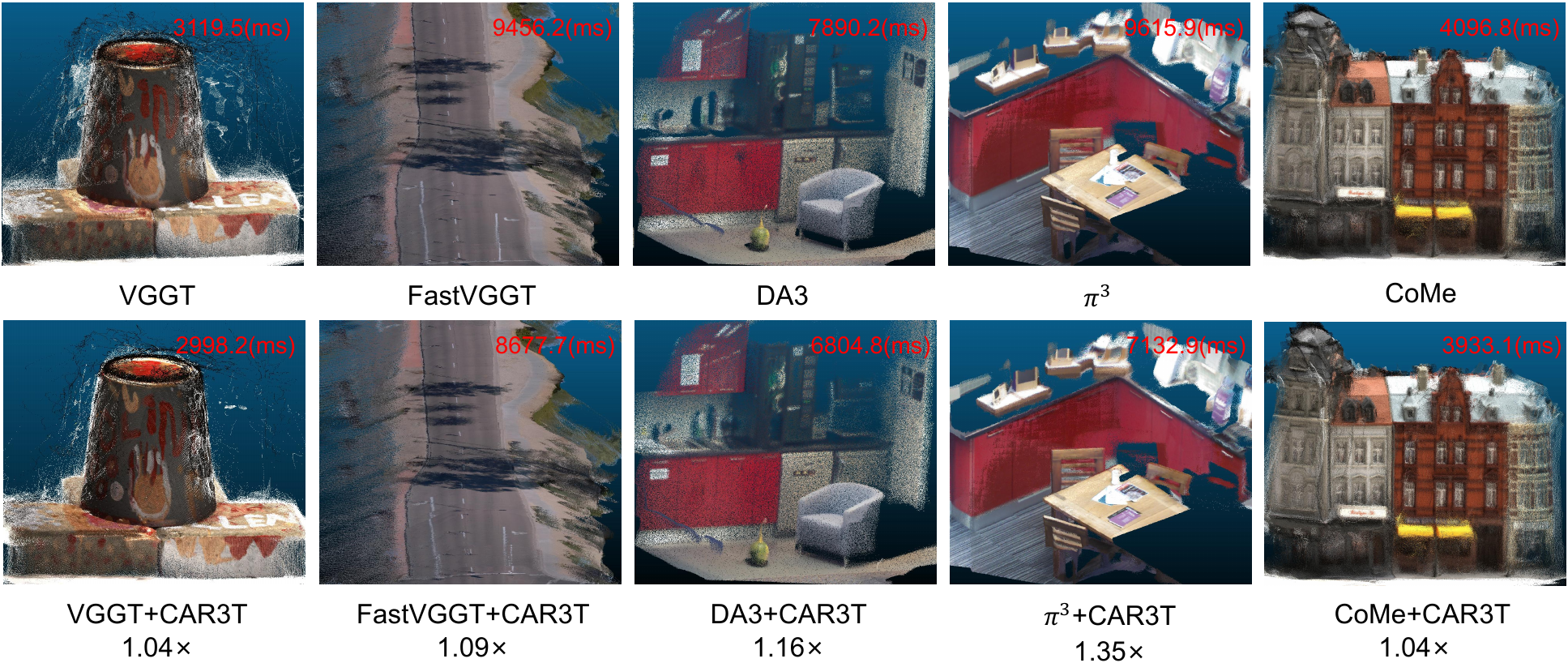}
    \caption{\textbf{Qualitative comparison of the 3D reconstruction results generated by CAR3T with those of various feedforward models.} By introducing layer reuse, CAR3T achieves visually comparable reconstruction quality.}
    \label{fig:4}
      \vspace{-3mm}
\end{figure*}
\noindent\textbf{Implementation Details.}
We use the images from ScanNet++ to perform offline calibration for each backbone network separately. 
We adopt four calibration tiers, each containing 50 scenes, with 38 scenes used for risk model fitting and 12 for budget calibration. 
Each scene contains up to 50, 150, 250, and 500 views, respectively. 
The number of views and tier configuration are set according to the requirements of the test inputs, and during online inference, the calibration tier is selected based on the number of input views.

\noindent\textbf{Models and Datasets.} 
We evaluate CAR3T on three models: DA3~\cite{lin2025da3}, $\pi^3$~\cite{wang2026pi3}, and VGGT~\cite{wang2025vggt}. Each pair of baselines shares the same pretrained weights, with only the feature reuse setting toggled. Reference view selection for each model is described in Appx.~\ref{app:reference_view}. The main experiments cover 86 scene settings across ETH3D~\cite{ETH3D}, 7Scenes~\cite{7scenes}, ScanNet++~\cite{scannet++}, DTU~\cite{DTU}, DTU-64~\cite{DTU}, Waymo~\cite{Waymo}, and vKITTI~\cite{vKITTI}. Inference efficiency is reported for all the above settings; camera pose results exclude vKITTI~\cite{vKITTI}; 3D reconstruction evaluation uses ETH3D~\cite{ETH3D}, 7Scenes~\cite{7scenes}, ScanNet++~\cite{scannet++}, and Waymo~\cite{Waymo}. Long-sequence experiments are conducted on 7Scenes~\cite{7scenes}, NRGBD~\cite{NRGBD}, and ScanNet++~\cite{scannet++}, while ablation experiments use a fixed VGGT~\cite{wang2025vggt} backbone.

\noindent\textbf{Inputs and Evaluation Metrics.}
In the main experiments and ablation studies, at most 100 frames are input for each scene: when fewer than 100 frames are available, all frames are retained, ensuring that the baselines and CAR3T use the same inputs. For long-sequence experiments, the input is fixed at 50, 150, 250, and 500 frames, with ScanNet++ results reported only up to 250 frames. We evaluate camera pose using AUC@$5^\circ$ and AUC@$30^\circ$, and 3D reconstruction using Acc., Comp., Overall, and F-score; metric details are provided in Sec.~\ref{app:metrics}. All experiments are conducted on an NVIDIA RTX PRO 6000 Blackwell GPU.

\subsection{Main Results}
\label{sec:5.2}

\begin{table*}[!t]
\caption{\textbf{Inference efficiency with up to 100 input frames per scene.} We report the forward pass time ($sec$/scene) and the speedup relative to the baseline (×). “Avg. frames” is the actual average input length and gray rows indicate the speedup.}
\label{tab:f100_speed}
\centering
\small
\setlength{\tabcolsep}{3pt}
\begin{tabular*}{\textwidth}{@{\extracolsep{\fill}}l*{7}{r}@{}}
\toprule
Method & ETH3D~\cite{ETH3D} & 7Scenes~\cite{7scenes} & ScanNet++ ~\cite{scannet++}& DTU~\cite{DTU} & DTU-64 ~\cite{DTU}& Waymo ~\cite{Waymo}& vKITTI ~\cite{vKITTI}\\
\midrule
Avg. frames & 34.3 & 100.0 & 100.0 & 49.0 & 55.9 & 100.0 & 100.0  \\
\midrule

DA3~\cite{lin2025da3} & 0.941 & 4.692 & 4.697 & 1.685 & 2.024 & 3.936 & 1.308  \\
DA3+CAR3T & 0.882 & 4.021 & 4.031  & 1.539 & 1.828 & 3.390 & 1.196 \\
\rlap{\smash{%
  \begingroup
  \setlength{\fboxsep}{0pt}%
  \colorbox[gray]{0.94}{%
    \makebox[\textwidth][l]{\strut}%
  }%
  \endgroup
}}%
\textbf{Speedup} ($\times$)$\uparrow$ & 1.067 & 1.167 & 1.165  & 1.095 & 1.107 & 1.161 & 1.094  \\
\addlinespace[3pt]
$\pi^3$~\cite{wang2026pi3} & 1.124 & 5.355 & 5.363  & 1.739 & 2.148 & 5.425 & 5.254  \\
$\pi^3$+CAR3T & 0.938 & 4.429 & 4.518  & 1.397 & 1.695 & 4.349 & 4.206  \\
\rlap{\smash{%
  \begingroup
  \setlength{\fboxsep}{0pt}%
  \colorbox[gray]{0.94}{%
    \makebox[\textwidth][l]{\strut}%
  }%
  \endgroup
}}%
\textbf{Speedup} ($\times$)$\uparrow$ & 1.199 & 1.209 & 1.187  & 1.245 & 1.267 & 1.247 & 1.249  \\
\addlinespace[3pt]
VGGT~\cite{wang2025vggt} & 1.037 & 5.484 & 5.504  & 1.896 & 2.302 & 4.604 & 1.391  \\
VGGT+CAR3T & 0.902 & 3.884 & 4.045  & 1.528 & 1.798 & 3.355 & 1.191  \\
\rlap{\smash{%
  \begingroup
  \setlength{\fboxsep}{0pt}%
  \colorbox[gray]{0.94}{%
    \makebox[\textwidth][l]{\strut}%
  }%
  \endgroup
}}%
\textbf{Speedup} ($\times$)$\uparrow$ & 1.150 & 1.412 & 1.361  & 1.241 & 1.281 & 1.372 & 1.167  \\
\bottomrule
\end{tabular*}
\end{table*}
We apply CAR3T to VGGT, DA3, and $\pi^3$ and evaluate them pairwise against their respective original models.

\noindent\textbf{Inference Efficiency.}
Table~\ref{tab:f100_speed} reports both the actual average number of input frames and the forward inference time. Across all scenes, CAR3T achieves speedups on all three models, demonstrating its ability to reduce inference overhead across architectures. 
The speedup gains are more pronounced on datasets with up to 100 input frames, such as 7Scenes, ScanNet++, and Waymo. 
These gains stem from computing global attention only for the reference view in the selected layers and reusing the resulting output across views. 
This reduces the core attention computation cost in these layers while preserving the full key-value context.

\noindent\textbf{Camera Pose Accuracy.}
Table~\ref{tab:f100_pose} reports pose accuracy under the strict AUC@$5^\circ$ metric. With $\pi^3$, CAR3T remains within 0.020 AUC of the baseline on five datasets and improves performance on ScanNet++. CAR3T also improves DA3 on 7Scenes, while results for DA3 and VGGT vary across datasets. These results illustrate the efficiency--accuracy trade-off of applying CAR3T to different backbones. Results under AUC@$30^\circ$ are provided in Table~\ref{tab:f100_pose_auc30} in the supplementary material.

\begin{table}[!htbp]
\caption{\textbf{Camera pose estimation accuracy with up to 100 input frames per scene.} We report AUC@$5^\circ$ ($\uparrow$) for each method.}
\label{tab:f100_pose}
\centering
\footnotesize
\setlength{\tabcolsep}{2pt}
\begin{tabular*}{\columnwidth}{@{\extracolsep{\fill}}l*{6}{r}@{}}
\toprule
 & \multicolumn{6}{c}{AUC@\(5^\circ\)\(\uparrow\)} \\
\cmidrule(l){2-7}
Dataset & \multicolumn{2}{c}{DA3~\cite{lin2025da3}} & \multicolumn{2}{c}{$\pi^3$~\cite{wang2026pi3}} & \multicolumn{2}{c}{VGGT~\cite{wang2025vggt}} \\
\cmidrule(lr){2-3}\cmidrule(lr){4-5}\cmidrule(l){6-7}
 & Base & CAR3T & Base & CAR3T & Base & CAR3T \\
\midrule
ETH3D ~\cite{ETH3D}    & 0.6686 & 0.4758 & 0.4715 & 0.4650 & 0.4347 & 0.3652 \\
7Scenes~\cite{7scenes}    & 0.4290 & 0.4433 & 0.4237 & 0.4068 & 0.3750 & 0.3461 \\
ScanNet++ ~\cite{scannet++}& 0.8877 & 0.6763 & 0.6952 & 0.7199 & 0.7208 & 0.6217 \\
DTU~\cite{DTU}       & 0.9423 & 0.7833 & 0.8321 & 0.8125 & 0.9924 & 0.8876 \\
DTU-64~\cite{DTU}    & 0.9410 & 0.8295 & 0.7362 & 0.7214 & 0.8087 & 0.7292 \\
Waymo ~\cite{Waymo}     & 0.9041 & 0.7258 & 0.5043 & 0.4865 & 0.7652 & 0.7465 \\
\bottomrule
\end{tabular*}
\end{table}

\noindent\textbf{3D Reconstruction Accuracy.}
Table~\ref{tab:f100_recon} shows that CAR3T reduces the Overall reconstruction error and improves the F-score of $\pi^3$ on ETH3D and 7Scenes. It also reduces VGGT's Overall error on 7Scenes. 
Results on ScanNet++ and Waymo are reported in Table~\ref{tab:f100_recon_supp} in the Appx.~\ref{appx:3dreconstruct}.

\begin{table}[!htbp]
\caption{\textbf{Pose-free 3D reconstruction quality on ETH3D and 7Scenes with up to 100 frames.} 
}
\label{tab:f100_recon}
\centering
\footnotesize
\setlength{\tabcolsep}{2pt}
\begin{tabular*}{\columnwidth}{@{\extracolsep{\fill}}ll*{4}{r}@{}}
\toprule
Model & Metric & \multicolumn{2}{c}{ETH3D~\cite{ETH3D}} & \multicolumn{2}{c}{7Scenes~\cite{7scenes}} \\
\cmidrule(lr){3-4}\cmidrule(lr){5-6}
 & & Base & CAR3T & Base & CAR3T \\
\midrule
DA3~\cite{lin2025da3} & Acc.$\downarrow$ & 0.2139 & 0.2560 & 0.1142 & 0.1086 \\
 & Comp.$\downarrow$ & 0.6192 & 0.7316 & 0.1331 & 0.1448 \\
 & Overall$\downarrow$ & 0.4166 & 0.4938 & 0.1237 & 0.1267 \\
 & F-score$\uparrow$ & 0.8202 & 0.6813 & 0.5141 & 0.5074 \\
\midrule
$\pi^3$~\cite{wang2026pi3} & Acc.$\downarrow$ & 0.2743 & 0.2572 & 0.1350 & 0.1271 \\
 & Comp.$\downarrow$ & 0.7680 & 0.7089 & 0.1105 & 0.1070 \\
 & Overall$\downarrow$ & 0.5212 & 0.4831 & 0.1228 & 0.1171 \\
 & F-score$\uparrow$ & 0.6943 & 0.7129 & 0.4152 & 0.4347 \\
\midrule
VGGT~\cite{wang2025vggt} & Acc.$\downarrow$ & 0.3074 & 0.3111 & 0.1487 & 0.1484 \\
 & Comp.$\downarrow$ & 0.8191 & 0.8648 & 0.1348 & 0.1301 \\
 & Overall$\downarrow$ & 0.5632 & 0.5880 & 0.1418 & 0.1393 \\
 & F-score$\uparrow$ & 0.6280 & 0.5963 & 0.4791 & 0.4817 \\
\bottomrule
\end{tabular*}
\end{table}

\noindent\textbf{Efficiency and Accuracy on Long Sequences.}
As shown in Table~\ref{tab:long_sequence}, CAR3T achieves inference acceleration across all tested configurations, and the acceleration gain generally increases as the input sequence length grows, indicating that feature reuse can further reduce inference overhead in long-sequence scenarios.
This trend stems from differences in computational complexity: the overhead of global attention grows quadratically with the number of views, whereas reusing attention outputs from reference-view attention can reduce the attention computation in selected layers to linear complexity. As a result, longer sequences can achieve greater computational savings from cross-view reuse of attention outputs.

For pose estimation, $\pi^3$ and VGGT generally maintain accuracy across sequence lengths, while DA3 remains close to the baseline in most settings. 
Reconstruction results are also largely stable, with improvements in several cases, although DA3 at 250 frames exhibits larger accuracy changes.
Overall, CAR3T demonstrates stable inference acceleration across different backbones and sequence lengths, while accuracy changes are jointly influenced by the backbone architecture and input sequence length.

\begin{table*}[t]
\caption{\textbf{Long-sequence performance.} Changes in pose and reconstruction accuracy relative to baseline models, and forward acceleration ratio of CAR3T across different sequence lengths. Reported metrics include $\Delta$AUC@30° (↑), $\Delta$Acc. (↓), $\Delta$Comp. (↓), $\Delta$F-score (↑), and acceleration ratio (×, ↑) ($\Delta$ = CAR3T - baseline). ScanNet++ is evaluated on up to 250 frames.}
\label{tab:long_sequence}
\centering
\scriptsize
\setlength{\tabcolsep}{1.5pt}
\resizebox{\textwidth}{!}{%
\begin{tabular}{@{}lr*{15}{r}@{}}
\toprule
Model & Frames & \multicolumn{5}{c}{7Scenes~\cite{7scenes}} & \multicolumn{5}{c}{NRGBD~\cite{NRGBD}} & \multicolumn{5}{c}{ScanNet++~\cite{scannet++}} \\
\cmidrule(lr){3-7}\cmidrule(lr){8-12}\cmidrule(l){13-17}
 & & \multicolumn{1}{c}{$\Delta$AUC@30$^\circ$} & \multicolumn{1}{c}{$\Delta$Acc.} & \multicolumn{1}{c}{$\Delta$Comp.} & \multicolumn{1}{c}{$\Delta$F-score} & \multicolumn{1}{c}{Spd.($\times$)$\uparrow$} & \multicolumn{1}{c}{$\Delta$AUC@30$^\circ$} & \multicolumn{1}{c}{$\Delta$Acc.} & \multicolumn{1}{c}{$\Delta$Comp.} & \multicolumn{1}{c}{$\Delta$F-score} & \multicolumn{1}{c}{Spd.($\times$)$\uparrow$} & \multicolumn{1}{c}{$\Delta$AUC@30$^\circ$} & \multicolumn{1}{c}{$\Delta$Acc.} & \multicolumn{1}{c}{$\Delta$Comp.} & \multicolumn{1}{c}{$\Delta$F-score} & \multicolumn{1}{c}{Spd.($\times$)$\uparrow$} \\
\midrule
DA3 ~\cite{lin2025da3}& 50 & +0.0062 & -0.0154 & +0.0068 & -0.0331 & 1.097 & -0.0077 & -0.0152 & +0.0046 & -0.0617 & 1.096 & -0.1000 & +0.0188 & +0.0199 & -0.1340 & 1.098 \\
 & 150 & +0.0033 & -0.0097 & +0.0193 & -0.0307 & 1.202 & -0.0138 & +0.0033 & +0.0048 & -0.0797 & 1.201 & -0.0406 & +0.0135 & +0.0132 & -0.1468 & 1.206 \\
 & 250 & -0.1931 & +0.2242 & +0.1255 & -0.3159 & 1.319 & -0.1701 & +0.1103 & +0.0860 & -0.5946 & 1.331 & -0.3275 & +0.1164 & +0.0968 & -0.5092 & 1.266 \\
 & 500 & -0.0019 & -0.0197 & +0.0193 & -0.0596 & 1.312 & -0.0156 & -0.0271 & +0.0061 & -0.0822 & 1.312 & -- & -- & -- & -- & -- \\
\midrule
$\pi^3$~\cite{wang2026pi3} & 50 & -0.0028 & -0.0011 & -0.0003 & +0.0098 & 1.256 & -0.0012 & -0.0037 & +0.0009 & -0.0153 & 1.256 & +0.0037 & -0.0008 & -0.0001 & +0.0120 & 1.238 \\
 & 150 & -0.0045 & -0.0053 & -0.0002 & +0.0042 & 1.248 & -0.0001 & -0.0192 & -0.0002 & +0.0119 & 1.250 & -0.0031 & +0.0055 & +0.0053 & -0.0674 & 1.212 \\
 & 250 & -0.0082 & -0.0141 & +0.0035 & +0.0134 & 1.684 & -0.0038 & +0.0168 & +0.0015 & -0.0287 & 1.669 & +0.0033 & +0.0014 & +0.0018 & -0.0264 & 1.576 \\
 & 500 & -0.0026 & -0.0176 & -0.0040 & +0.0387 & 1.553 & -0.0014 & -0.0011 & +0.0002 & -0.0154 & 1.537 & -- & -- & -- & -- & -- \\
\midrule
VGGT~\cite{wang2025vggt} & 50 & -0.0117 & +0.0015 & -0.0004 & +0.0008 & 1.271 & -0.0098 & +0.0071 & +0.0077 & -0.0684 & 1.262 & -0.0269 & +0.0056 & +0.0041 & -0.0458 & 1.225 \\
 & 150 & -0.0104 & -0.0185 & -0.0250 & +0.0067 & 1.552 & -0.0180 & +0.0088 & +0.0067 & -0.0634 & 1.547 & -0.0224 & +0.0055 & +0.0050 & -0.0609 & 1.463 \\
 & 250 & -0.0142 & +0.0141 & -0.0016 & -0.0012 & 1.726 & -0.0256 & +0.0236 & +0.0139 & -0.1091 & 1.703 & -0.0272 & +0.0049 & +0.0066 & -0.0642 & 1.555 \\
 & 500 & -0.0079 & +0.0432 & -0.0021 & -0.0201 & 1.733 & -0.0881 & +0.2380 & +0.1010 & -0.0947 & 1.748 & -- & -- & -- & -- & -- \\
\bottomrule
\end{tabular}
}
\end{table*}

\subsection{Compatibility with Existing Acceleration Methods}
Table~\ref{tab:come+fastvggt} evaluates CAR3T applied to VGGT and the acceleration methods FastVGGT~\cite{shen2025fastvggt} and CoMe~\cite{chen2026come}. CAR3T provides additional acceleration for both methods across all three datasets, demonstrating its compatibility with existing acceleration approaches.
This is because token merging shortens the sequence length but does not explicitly eliminate the similarity between attention distributions across different views. CAR3T leverages this residual redundancy by reusing the attention outputs of the reference view in the selected layer, thereby further reducing computational cost on top of token merging.
Fig.~\ref{fig:4} visualizes reconstructions from 50--200 input frames. CAR3T preserves the major geometric structures and scene layout of the original models while reducing inference overhead.

\begin{table*}[t]
\centering
\caption{\textbf{Accuracy and inference efficiency of VGGT~\cite{wang2025vggt}, FastVGGT~\cite{shen2025fastvggt}, and CoMe~\cite{chen2026come} with and without CAR3T on three datasets.} Speedup is measured relative to VGGT. The merging threshold for CoMe is 0.5 and the merge ratio of FastVGGT is 0.9. }
\setlength{\tabcolsep}{4pt}
\label{tab:come+fastvggt}
\renewcommand{\arraystretch}{1.16}
\resizebox{\textwidth}{!}{%
\begin{tabular}{l*{15}{c}}
\toprule
Method
& \multicolumn{5}{c}{7Scenes~\cite{7scenes}}
& \multicolumn{5}{c}{ScanNet++~\cite{scannet++}}
& \multicolumn{5}{c}{Waymo~\cite{Waymo}} \\
\cmidrule(lr){2-6}\cmidrule(lr){7-11}\cmidrule(lr){12-16}
& AUC@\(30^\circ\) $\uparrow$ & ACC. $\downarrow$ & Comp. $\downarrow$ & F-Score $\uparrow$ & Spd.($\times$) $\uparrow$
& AUC@\(30^\circ\) $\uparrow$ & ACC. $\downarrow$ & Comp. $\downarrow$ & F-Score $\uparrow$ & Spd.($\times$) $\uparrow$
& AUC@\(30^\circ\) $\uparrow$ & ACC. $\downarrow$ & Comp. $\downarrow$ & F-Score $\uparrow$ & Spd.($\times$) $\uparrow$ \\
\midrule
VGGT~\cite{wang2025vggt}
& 84.083 & 0.1847 & 0.1239 & 47.020 & 1.000
& 93.606 & 0.0486 & 0.0829 & 68.201 & 1.000
& 93.540 & 0.5587 & 0.5126 & 43.194 & 1.000 \\
\rowcolor[gray]{0.95} VGGT +CAR3T
& 82.557 & 0.2031 & 0.1218 & 46.497 & 1.776
& 91.417 & 0.0541 & 0.0896 & 62.867 & 1.577
& 93.185 & 0.6813 & 0.6497 & 41.247 & 1.604 \\
\addlinespace[4pt]
FastVGGT~\cite{shen2025fastvggt}
& 83.058 & 0.1535 & 0.1360 & 47.620 & 2.488
& 90.565 & 0.0528 & 0.0901 & 59.656 & 2.373
& 93.297 & 0.6892 & 0.6360 & 36.894 & 2.037 \\
\rowcolor[gray]{0.95} FastVGGT +CAR3T
& 67.535 & 0.2287 & 0.1422 & 38.813 & 3.046
& 82.864 & 0.0712 & 0.1052 & 46.669 & 2.801
& 93.079 & 0.7506 & 0.7307 & 29.489 & 2.149 \\
\addlinespace[4pt]
CoMe~\cite{chen2026come}
& 79.905 & 0.1947 & 0.1921 & 35.147 & 2.058
& 69.543 & 0.0991 & 0.1315 & 42.892 & 2.010
& 85.712 & 1.5746 & 2.2182 & 24.428 & 1.506 \\
\rowcolor[gray]{0.95} CoMe +CAR3T
& 79.912 & 0.1931 & 0.1904 & 36.332 & 2.151
& 69.636 & 0.1001 & 0.1322 & 42.263 & 2.119
& 85.539 & 1.7505 & 2.6966 & 23.431 & 1.887 \\
\bottomrule
\end{tabular}%
}
\end{table*}


\subsection{Ablation Studies}
\label{sec:5.4}

To analyze the roles of the reuse strategy and budget selection in CAR3T, we conduct ablation studies with the VGGT backbone on several datasets. 

\noindent\textbf{Reuse Strategy.}
Table~\ref{tab:vggt_reuse_strategy} compares full-layer reuse, fixed 12-layer and 14-layer configurations constructed from the low-JSD layers in Fig.~\ref{fig:3}, and CAR3T. Although full-layer reuse achieves the highest speedup, it causes a substantial drop in pose and reconstruction accuracy. 
The two fixed low-JSD configurations also achieve speedup, but their accuracy is significantly lower than that of CAR3T. In particular, compared with the fixed 12-layer configuration, CAR3T achieves nearly the same average number of reused layers while attaining higher AUC@$30^\circ$ and F-score and lower Overall distance. 
This indicates that low JSD can be used to filter candidate layers, but relying solely on a pre-determined fixed set of layers is still insufficient to guarantee prediction quality across different inputs; CAR3T’s input-adaptive decision improves accuracy preservation under a similar reuse scale.

\begin{table}[H]
\caption{\textbf{Speed and accuracy comparison of different global attention reuse strategies on VGGT.} 
}
\label{tab:vggt_reuse_strategy}
\centering
\footnotesize
\setlength{\tabcolsep}{2pt}
\begin{tabular*}{\columnwidth}{@{\extracolsep{\fill}}lrrrrr@{}}
\toprule
Variant & Layers & Spd.($\times$)$\uparrow$ & AUC@\(30^\circ\)\(\uparrow\) & F-score$\uparrow$ & Overall$\downarrow$ \\
\midrule
Baseline & 0.00 & 1.000 & 0.8725 & 0.5825 & 0.2230 \\
All layers & 24.00 & 2.064 & 0.0199 & 0.0249 & 2.6524 \\
Static (12) & 12.00 & 1.363 & 0.7988 & 0.4676 & 0.2512 \\
Static (14) & 14.00 & 1.438 & 0.7940 & 0.4626 & 0.2564 \\
\midrule
\textbf{CAR3T} & 12.19 & 1.317 & 0.8540 & 0.5577 & 0.2263 \\
\bottomrule
\end{tabular*}
\end{table}

\noindent\textbf{Budget Sensitivity.}
Table~\ref{tab:vggt_budget_sensitivity} further examines the impact of the quality budget. When the median of the candidate Pareto front risk is used instead of the inflection-point budget, both the number of reused layers and the speedup decrease, although accuracy improves slightly. This suggests that the default inflection-point selection can obtain more substantial efficiency gains at a limited accuracy cost. As the budget increases from $0.5\times$ to $1.25\times$, the speedup generally increases, while AUC@$30^\circ$ and F-score gradually decrease; further increasing it to $1.5\times$ does not yield additional speedup. The default budget therefore lies in a relatively reasonable trade-off region between efficiency and accuracy: further relaxing the budget yields only limited speed gains while increasing accuracy loss.
These two sets of experiments show that the benefits of CAR3T come from the combination of adaptive reuse-layer selection and budget constraints: the former demonstrates the necessity of dynamic layer selection, while the latter controls the trade-off between speedup gains and output quality.

\begin{table}[H]
\caption{\textbf{Impact of reuse budget settings on VGGT inference speed and accuracy.} No knee uses the median risk of the candidate Pareto front instead of the knee-point budget. Budget multipliers scale the default calibrated budget used by CAR3T.}
\label{tab:vggt_budget_sensitivity}
\centering
\footnotesize
\setlength{\tabcolsep}{2pt}
\begin{tabular*}{\columnwidth}{@{\extracolsep{\fill}}lrrrrr@{}}
\toprule
Variant & Layers & Spd.($\times$)$\uparrow$ & AUC@\(30^\circ\)\(\uparrow\) & F-score$\uparrow$ & Overall$\downarrow$ \\
\midrule
No knee & 8.88 & 1.200 & 0.8621 & 0.5718 & 0.2234 \\
Budget $0.5	\times$ & 7.75 & 1.181 & 0.8650 & 0.5700 & 0.2259 \\
Budget $0.75\times$ & 10.26 & 1.263 & 0.8611 & 0.5754 & 0.2225 \\
Budget $1.25\times$ & 13.25 & 1.328 & 0.8527 & 0.5551 & 0.2292 \\
Budget $1.5\times$ & 12.65 & 1.210 & 0.8469 & 0.5417 & 0.2247 \\
\midrule
\textbf{CAR3T} & 12.19 & 1.317 & 0.8540 & 0.5577 & 0.2263 \\
\bottomrule
\end{tabular*}
  \vspace{-3mm}
\end{table}

\section{Conclusion}
\label{sec:conclusion}

We propose CAR3T, a training-free framework that accelerates multi-view 3D reconstruction by reusing reference-view attention outputs. It combines offline risk calibration with online query drift estimation and layer selection to adapt to layer- and input-dependent attention redundancy while controlling reuse error. Despite the additional overhead of these online decisions, experiments show that CAR3T reduces inference time while largely preserving reconstruction quality.


\clearpage
\bibliographystyle{unsrt}  
\bibliography{references} 


\clearpage
\appendix

\clearpage
\appendix
\section*{Appendix}
\section*{Contents}
\noindent\hyperref[sec:Preliminary]{A. Preliminary} \dotfill \pageref{sec:Preliminary}\\
\hyperref[app:implementation]{B. Implementation Details} \dotfill \pageref{app:implementation}\\
\hyperref[app:additional_experiments]{C. Additional Experiments} \dotfill \pageref{app:additional_experiments}\\
\hyperref[app:additional_visualizations]{D. Additional Visualizations} \dotfill \pageref{app:additional_visualizations}

\section{Preliminary}
\label{sec:Preliminary}

VGGT, $\pi^3$, and DA3 are representative feed-forward geometric models that achieve cross-view information aggregation through global attention \cite{wang2025vggt,wang2026pi3,lin2025da3}. Although their specific architectures differ, the core computation of global attention in all three can be expressed as scaled dot-product attention between view-specific Queries and a global Key-Value (KV) set.

Given an input set $\mathcal{I} = \{I_i\}_{i=1}^{M}$ composed of $M$ views, the model encodes each view into $P$ tokens, including image patch tokens and special tokens. Denote the features belonging to view $i$ at layer $\ell$ as
\begin{equation}
\mathbf{X}_i^\ell
\in
\mathbb{R}^{B \times P \times C},
\label{eq:view_features}
\end{equation}
where $B$ and $C$ denote the batch size and feature dimension, respectively.  
Let $\mathbf{Q}_i^\ell$ represent the Query of view $i$, and concatenate all views' Keys and Values into
\begin{equation}
\mathbf{K}^\ell
=
[\mathbf{K}_1^\ell;\ldots;\mathbf{K}_M^\ell],
\qquad
\mathbf{V}^\ell
=
[\mathbf{V}_1^\ell;\ldots;\mathbf{V}_M^\ell].
\label{eq:global_kv}
\end{equation}
The global attention output corresponding to view $i$ is
\begin{equation}
\mathbf{O}_i^\ell
=
\operatorname{Softmax}
\left(
\frac{
\mathbf{Q}_i^\ell
(\mathbf{K}^\ell)^{\top}
}{\sqrt{d}}
\right)
\mathbf{V}^\ell,
\label{eq:view_global_attention}
\end{equation}
Here, $d$ denotes the feature dimension of a single attention head \cite{vaswani2017attention}.  
$\mathbf{O}_i^\ell$ represents the slice in the global attention output sequence corresponding to view $i$, and the complete output can be expressed as $\{\mathbf{O}_i^\ell\}_{i=1}^{M}$.

Full global attention requires computing interactions between all $MP$ queries and $MP$ keys.  
Ignoring batch dimensions and the number of attention heads, its core computational complexity is $\mathcal{O}(M^2P^2d)$, which grows quadratically with the number of input views \cite{vaswani2017attention,dao2022flashattention}.

\section{Implementation Details}
\label{app:implementation}
\subsection{Cross-View Attention JSD}
\label{app:jsd}
For layer $\ell$ and attention head $h$, let $A_{r,h,n}^{(\ell)}$ and $A_{v,h,n}^{(\ell)}$ denote the global attention distributions of query tokens at the same intra-view position $n$ in the reference view $r$ and a non-reference view $v$, respectively. Each distribution is defined over all $MP$ key tokens, where $M$ is the number of views and $P$ is the number of key tokens per view. Their Jensen--Shannon divergence is
\begin{equation}
\begin{aligned}
\operatorname{JSD}_{v,h,n}^{(\ell)}
={}&\frac{1}{2}D_{\mathrm{KL}}\left(A_{r,h,n}^{(\ell)}\,\|\,\bar{A}_{v,h,n}^{(\ell)}\right)\\
&+\frac{1}{2}D_{\mathrm{KL}}\left(A_{v,h,n}^{(\ell)}\,\|\,\bar{A}_{v,h,n}^{(\ell)}\right),
\end{aligned}
\end{equation}
where $\bar{A}_{v,h,n}^{(\ell)}=\frac{1}{2}(A_{r,h,n}^{(\ell)}+A_{v,h,n}^{(\ell)})$. Averaging over all non-reference views and the $P_q$ query positions per view gives the score for head $h$ at layer $\ell$:
\begin{equation}
J_h^{(\ell)}=
\frac{1}{(M-1)P_q}
\sum_{\substack{v=1\\v\neq r}}^{M}
\sum_{n=1}^{P_q}\operatorname{JSD}_{v,h,n}^{(\ell)}.
\end{equation}
The plotted mean and interquartile range are computed across attention heads.

\subsection{Reference View Selection}
\label{app:reference_view}
VGGT uses the first frame of the input sequence as the reference view. DA3 retains its original reference view selection strategy (default: \texttt{saddle\_balanced}) and moves the selected frame to the first position.
Before the first global attention block, $\pi^3$ constructs a normalized descriptor $e_i$ for each view from its register-token queries and selects the view with the smallest mean cosine distance to the other views:
\begin{equation}
r=
\underset{i\in\{1,\ldots,M\}}{\arg\min}
\frac{1}{M-1}
\sum_{\substack{j=1\\j\neq i}}^{M}
\left(1-e_i^\top e_j\right).
\end{equation}

\section{Additional Experiments}
\label{app:additional_experiments}
\subsection{Evaluation Metrics}
\label{app:metrics}
Camera pose is evaluated using AUC@$5^\circ$ and AUC@$30^\circ$. For 3D reconstruction, Acc. and Comp. measure the average nearest-neighbor distances from predicted points to ground-truth points and from ground-truth points to predicted points, respectively; $\mathrm{Overall}=(\mathrm{Acc.}+\mathrm{Comp.})/2$. The F-score is the harmonic mean of precision and recall, with the distance threshold set to 0.05 m or 0.25 m depending on the dataset. In the long-sequence table, all $\Delta$ metrics are defined as the CAR3T result minus the corresponding baseline result.

\subsection{Additional Reconstruction Results}
  \label{appx:3dreconstruct}
Table~\ref{tab:f100_recon_supp} complements Table~\ref{tab:f100_recon} with results on ScanNet++ and Waymo.

\begin{table}[!htbp]
\caption{\textbf{Pose-free 3D reconstruction quality on ScanNet++ and Waymo with up to 100 frames.} We report Acc., Comp., and Overall error (in meters, $\downarrow$), and F-score ($\uparrow$).}
\label{tab:f100_recon_supp}
\centering
\footnotesize
\setlength{\tabcolsep}{2pt}
\begin{tabular*}{\columnwidth}{@{\extracolsep{\fill}}ll*{4}{r}@{}}
\toprule
Model & Metric & \multicolumn{2}{c}{ScanNet++~\cite{scannet++}} & \multicolumn{2}{c}{Waymo~\cite{Waymo}} \\
\cmidrule(lr){3-4}\cmidrule(lr){5-6}
 & & Base & CAR3T & Base & CAR3T \\
\midrule
DA3~\cite{lin2025da3} & Acc.$\downarrow$ & 0.0321 & 0.0450 & 1.3344 & 1.1104 \\
 & Comp.$\downarrow$ & 0.1011 & 0.1134 & 0.9976 & 0.8536 \\
 & Overall$\downarrow$ & 0.0666 & 0.0792 & 1.1660 & 0.9820 \\
 & F-score$\uparrow$ & 0.7876 & 0.6531 & 0.4367 & 0.2103 \\
\midrule
$\pi^3$~\cite{wang2026pi3} & Acc.$\downarrow$ & 0.0466 & 0.0464 & 3.6148 & 3.4673 \\
 & Comp.$\downarrow$ & 0.1142 & 0.1140 & 2.4804 & 2.5011 \\
 & Overall$\downarrow$ & 0.0804 & 0.0802 & 3.0476 & 2.9842 \\
 & F-score$\uparrow$ & 0.6316 & 0.6373 & 0.1639 & 0.1444 \\
\midrule
VGGT~\cite{wang2025vggt} & Acc.$\downarrow$ & 0.0442 & 0.0478 & 0.7861 & 0.8322 \\
 & Comp.$\downarrow$ & 0.1140 & 0.1171 & 0.8053 & 0.9119 \\
 & Overall$\downarrow$ & 0.0791 & 0.0825 & 0.7957 & 0.8721 \\
 & F-score$\uparrow$ & 0.6757 & 0.6402 & 0.3595 & 0.3960 \\
\bottomrule
\end{tabular*}
\end{table}

\subsection{Additional Camera Pose Results}
Table~\ref{tab:f100_pose_auc30} reports AUC@$30^\circ$ for the experiments in Table~\ref{tab:f100_pose}. CAR3T maintains performance close to the original models across the datasets, with improvements for $\pi^3$ on ETH3D and ScanNet++ and for DA3 on 7Scenes.

\begin{table}[t]
\caption{\textbf{Camera pose estimation accuracy with up to 100 input frames per scene.} We report AUC@$30^\circ$ ($\uparrow$).}
\label{tab:f100_pose_auc30}
\centering
\footnotesize
\setlength{\tabcolsep}{2pt}
\begin{tabular*}{\columnwidth}{@{\extracolsep{\fill}}l*{6}{r}@{}}
\toprule
 & \multicolumn{6}{c}{AUC@\(30^\circ\)\(\uparrow\)} \\
\cmidrule(l){2-7}
Dataset & \multicolumn{2}{c}{DA3~\cite{lin2025da3}} & \multicolumn{2}{c}{$\pi^3$~\cite{wang2026pi3}} & \multicolumn{2}{c}{VGGT~\cite{wang2025vggt}} \\
\cmidrule(lr){2-3}\cmidrule(lr){4-5}\cmidrule(l){6-7}
 & Base & CAR3T & Base & CAR3T & Base & CAR3T \\
\midrule
ETH3D~\cite{ETH3D}     & 0.9229 & 0.8786 & 0.8556 & 0.8565 & 0.8181 & 0.7851 \\
7Scenes ~\cite{7scenes}  & 0.8651 & 0.8729 & 0.8614 & 0.8568 & 0.8474 & 0.8373 \\
ScanNet++ ~\cite{scannet++} & 0.9797 & 0.9404 & 0.9376 & 0.9469 & 0.9438 & 0.9247 \\
DTU ~\cite{DTU}       & 0.9904 & 0.9609 & 0.9717 & 0.9682 & 0.9987 & 0.9813 \\
DTU-64~\cite{DTU}   & 0.9901 & 0.9702 & 0.9486 & 0.9473 & 0.8526 & 0.8378 \\
Waymo ~\cite{Waymo}    & 0.9682 & 0.9324 & 0.8765 & 0.8748 & 0.9397 & 0.9362 \\
\bottomrule
\end{tabular*}
\end{table}

\section{Additional Visualizations}
\label{app:additional_visualizations}

\begin{figure*}[t]
    \centering
    \includegraphics[width=1\textwidth]{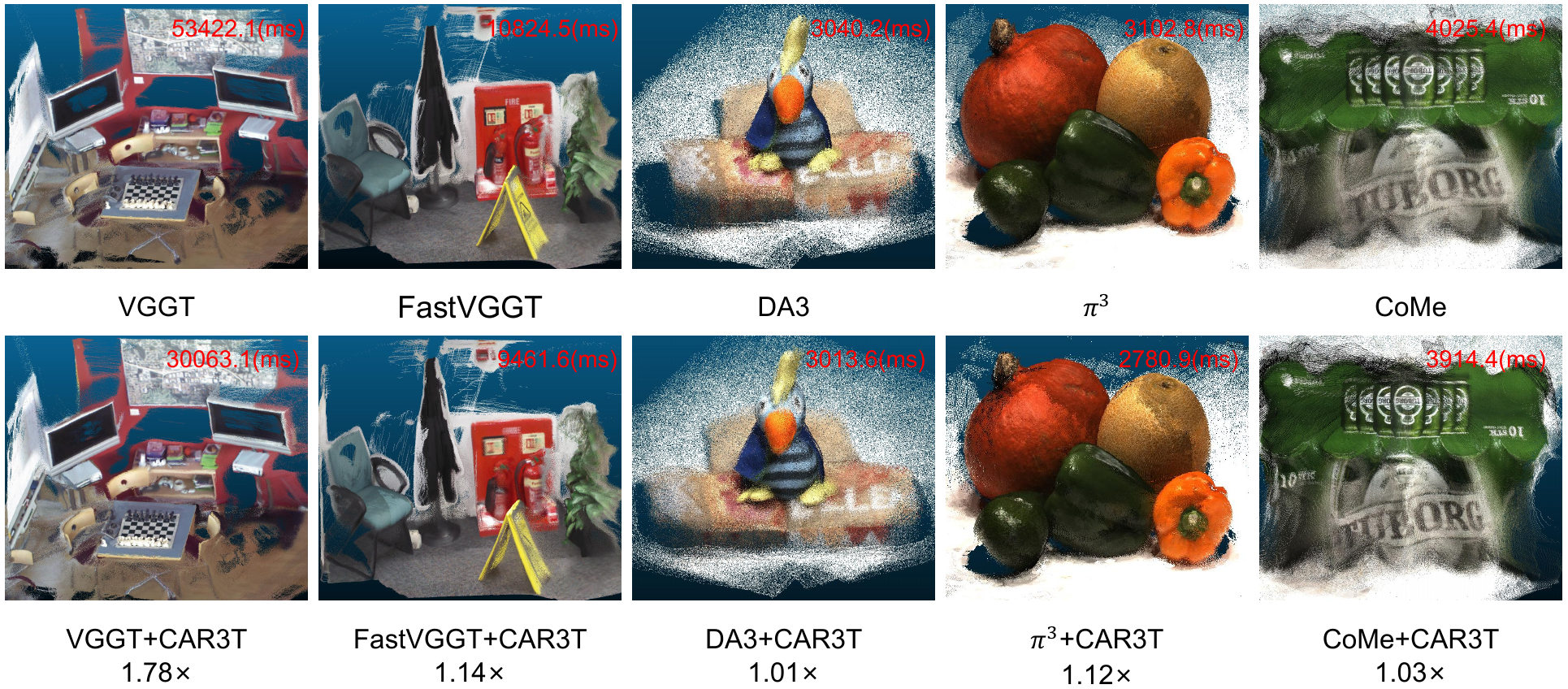}
    \caption{\textbf{Qualitative comparison of the 3D reconstruction results generated by CAR3T with those of various feedforward models.} By introducing layer reuse, CAR3T achieves visually comparable reconstruction quality.}
    \label{fig:suppl}
\end{figure*}

\end{document}